\documentclass[11pt]{article}

\usepackage{colacl,amsmath,rotate,psfig}

\author{Detlef Prescher \and Stefan Riezler \and Mats Rooth \\
Institut f\"ur Maschinelle Sprachverarbeitung \\
Universit\"at Stuttgart, Germany}

\title{Using a Probabilistic Class-Based Lexicon \\
 for Lexical Ambiguity Resolution} 

\begin{document}

\maketitle

\begin{abstract}
This paper presents the use of probabilistic class-based lexica
for disambiguation in target-word selection.
Our method employs minimal but precise contextual information for
disambiguation. That is, only information provided by the 
target-verb, enriched by the condensed information of a probabilistic
class-based lexicon, is used. Induction of classes and fine-tuning to
verbal arguments is done in an unsupervised manner by EM-based
clustering techniques. The method shows promising results in an
evaluation on real-world translations. 
\end{abstract}

\section{Introduction}
\label{SecIntro}

Disambiguation of lexical ambiguities in naturally occuring free text
is considered a hard task for computational linguistics. For instance, word sense disambiguation is concerned with the problem of
assigning sense labels to occurrences of an ambiguous word. Resolving
such ambiguities is useful in constraining semantic interpretation. 
A related task is target-word disambiguation in machine translation.
Here a decision has to be made which of a set of alternative
target-language words is the most appropriate translation
of a source-language word. A solution to this disambiguation problem
is directly applicable in a machine translation system which is able
to propose the translation alternatives.
A further problem is the resolution of attachment ambiguities in
syntactic parsing. Here the decision of verb versus argument
attachment of noun phrases, or the choice for verb phrase versus noun
phrase attachment of prepositional phrases can build upon a resolution
of the related lexical ambiguities.

Statistical approaches have been applied successfully to these
problems. The great advantage of statistical methods over
symbolic-linguistic methods has been deemed to be their effective
exploitation of minimal linguistic knowledge. However, the best
performing statistical approaches to lexical ambiguity resolution themselves rely on complex information sources
such as ``lemmas, inflected forms, parts of speech and arbitrary word
classes [\ldots] local and distant collocations, trigram sequences,
and predicate argument association''~(Yarowsky \shortcite{Yarowsky:95},
p. 190) or large context-windows up to 1000 neighboring words~\cite{Schuetze:92}. Unfortunately, in many applications such
information is not readily available. For instance, in
incremental machine translation, it may be desirable to decide for the
most probable translation of the arguments of a verb with only the
translation of the verb as information source but no large window of
surrounding translations available. In parsing, the 
attachment of a nominal head may have to be resolved with only 
information about the semantic roles of the verb but no other predicate
argument associations at hand.
 
\begin{figure*}[ht]
\begin{center}
{\tiny
\setlength{\tabcolsep}{3pt}
\begin{tabular}
{|l|r|rrrrrrrrrrrrrrrrrrrrrrrrrrrrrr|} \hline
\begin{tabular}{c}
     {\bf Class 19} \\
                                 \\
     PROB 0.0235\\
                                 \\
\end{tabular}
&  & \rotate[l]{0.0250} 
 & \rotate[l]{0.0211} 
 & \rotate[l]{0.0125} 
 & \rotate[l]{0.0111} 
 & \rotate[l]{0.0106} 
 & \rotate[l]{0.0096} 
 & \rotate[l]{0.0085} 
 & \rotate[l]{0.0084} 
 & \rotate[l]{0.0081} 
 & \rotate[l]{0.0079} 
 & \rotate[l]{0.0076} 
 & \rotate[l]{0.0068} 
 & \rotate[l]{0.0067} 
 & \rotate[l]{0.0061} 
 & \rotate[l]{0.0059} 
 & \rotate[l]{0.0056} 
 & \rotate[l]{0.0053} 
 & \rotate[l]{0.0050} 
 & \rotate[l]{0.0049} 
 & \rotate[l]{0.0049} 
 & \rotate[l]{0.0048} 
 & \rotate[l]{0.0048} 
 & \rotate[l]{0.0047} 
 & \rotate[l]{0.0047} 
 & \rotate[l]{0.0046} 
 & \rotate[l]{0.0046} 
 & \rotate[l]{0.0045} 
 & \rotate[l]{0.0044} 
 & \rotate[l]{0.0043} 
 & \rotate[l]{0.0043} 
 \\ 
 \hline
&  & \rotate[l]{room} 
 & \rotate[l]{area} 
 & \rotate[l]{world} 
 & \rotate[l]{meeting} 
 & \rotate[l]{range} 
 & \rotate[l]{school} 
 & \rotate[l]{service} 
 & \rotate[l]{building} 
 & \rotate[l]{road} 
 & \rotate[l]{street} 
 & \rotate[l]{market} 
 & \rotate[l]{course} 
 & \rotate[l]{doctor} 
 & \rotate[l]{place} 
 & \rotate[l]{part} 
 & \rotate[l]{mind} 
 & \rotate[l]{class} 
 & \rotate[l]{scene} 
 & \rotate[l]{path} 
 & \rotate[l]{group} 
 & \rotate[l]{work} 
 & \rotate[l]{gray} 
 & \rotate[l]{programme} 
 & \rotate[l]{line} 
 & \rotate[l]{life} 
 & \rotate[l]{garden} 
 & \rotate[l]{body} 
 & \rotate[l]{miles} 
 & \rotate[l]{system} 
 & \rotate[l]{period} 
 \\ 
 \hline
0.0629 & enter.aso:o & $\bullet$ & $\bullet$ & $\bullet$ &  & $\bullet$ & $\bullet$ & $\bullet$ & $\bullet$ & $\bullet$ & $\bullet$ & $\bullet$ &  & $\bullet$ & $\bullet$ & $\bullet$ & $\bullet$ & $\bullet$ & $\bullet$ & $\bullet$ & $\bullet$ & $\bullet$ & $\bullet$ & $\bullet$ &  & $\bullet$ & $\bullet$ & $\bullet$ &  & $\bullet$ & $\bullet$\\
0.0386 & cover.aso:o &  & $\bullet$ & $\bullet$ & $\bullet$ & $\bullet$ & $\bullet$ & $\bullet$ &  & $\bullet$ & $\bullet$ & $\bullet$ & $\bullet$ & $\bullet$ & $\bullet$ & $\bullet$ &  &  & $\bullet$ &  & $\bullet$ & $\bullet$ & $\bullet$ & $\bullet$ & $\bullet$ &  &  & $\bullet$ & $\bullet$ &  & $\bullet$\\
0.0321 & call.aso:o & $\bullet$ & $\bullet$ & $\bullet$ & $\bullet$ &  & $\bullet$ & $\bullet$ &  & $\bullet$ & $\bullet$ &  &  &  & $\bullet$ & $\bullet$ & $\bullet$ & $\bullet$ & $\bullet$ & $\bullet$ & $\bullet$ & $\bullet$ & $\bullet$ & $\bullet$ & $\bullet$ & $\bullet$ &  &  &  & $\bullet$ & $\bullet$\\
0.0236 & include.aso:o & $\bullet$ & $\bullet$ & $\bullet$ & $\bullet$ & $\bullet$ & $\bullet$ & $\bullet$ & $\bullet$ &  & $\bullet$ &  & $\bullet$ & $\bullet$ & $\bullet$ & $\bullet$ & $\bullet$ & $\bullet$ &  &  & $\bullet$ & $\bullet$ & $\bullet$ & $\bullet$ & $\bullet$ & $\bullet$ & $\bullet$ & $\bullet$ & $\bullet$ & $\bullet$ & $\bullet$\\
0.0226 & run.aso:o & $\bullet$ &  & $\bullet$ &  &  & $\bullet$ & $\bullet$ &  &  &  &  & $\bullet$ & $\bullet$ & $\bullet$ & $\bullet$ &  & $\bullet$ &  & $\bullet$ & $\bullet$ & $\bullet$ &  & $\bullet$ & $\bullet$ & $\bullet$ &  &  & $\bullet$ & $\bullet$ & \\
0.0214 & attend.aso:o &  &  &  & $\bullet$ &  & $\bullet$ & $\bullet$ &  & $\bullet$ &  & $\bullet$ &  & $\bullet$ & $\bullet$ & $\bullet$ &  & $\bullet$ & $\bullet$ &  & $\bullet$ & $\bullet$ &  & $\bullet$ &  &  &  &  &  &  & \\
0.0173 & cross.aso:o & $\bullet$ & $\bullet$ &  &  & $\bullet$ &  &  & $\bullet$ & $\bullet$ & $\bullet$ &  & $\bullet$ &  &  & $\bullet$ & $\bullet$ &  &  & $\bullet$ &  &  & $\bullet$ &  & $\bullet$ &  & $\bullet$ & $\bullet$ &  &  & \\
0.0136 & dominate.aso:o & $\bullet$ & $\bullet$ & $\bullet$ & $\bullet$ &  &  & $\bullet$ & $\bullet$ & $\bullet$ & $\bullet$ & $\bullet$ &  &  & $\bullet$ &  &  & $\bullet$ & $\bullet$ &  &  & $\bullet$ & $\bullet$ &  &  & $\bullet$ &  & $\bullet$ &  & $\bullet$ & $\bullet$\\
0.0132 & have.aso:s & $\bullet$ &  & $\bullet$ & $\bullet$ & $\bullet$ & $\bullet$ & $\bullet$ & $\bullet$ & $\bullet$ & $\bullet$ & $\bullet$ & $\bullet$ &  & $\bullet$ & $\bullet$ & $\bullet$ & $\bullet$ & $\bullet$ & $\bullet$ &  & $\bullet$ & $\bullet$ & $\bullet$ & $\bullet$ & $\bullet$ & $\bullet$ & $\bullet$ & $\bullet$ & $\bullet$ & $\bullet$\\
0.0126 & attract.aso:s &  & $\bullet$ &  & $\bullet$ &  & $\bullet$ & $\bullet$ &  & $\bullet$ &  &  & $\bullet$ &  & $\bullet$ & $\bullet$ &  &  & $\bullet$ &  & $\bullet$ & $\bullet$ &  & $\bullet$ &  &  &  &  & $\bullet$ & $\bullet$ & $\bullet$\\
0.0124 & occupy.aso:o & $\bullet$ & $\bullet$ & $\bullet$ &  & $\bullet$ &  &  &  &  & $\bullet$ &  &  & $\bullet$ & $\bullet$ & $\bullet$ & $\bullet$ &  &  &  &  &  & $\bullet$ &  &  &  &  & $\bullet$ & $\bullet$ &  & \\
0.0115 & include.aso:s & $\bullet$ & $\bullet$ &  & $\bullet$ & $\bullet$ & $\bullet$ & $\bullet$ & $\bullet$ & $\bullet$ &  & $\bullet$ & $\bullet$ & $\bullet$ & $\bullet$ & $\bullet$ & $\bullet$ & $\bullet$ & $\bullet$ & $\bullet$ &  & $\bullet$ & $\bullet$ &  & $\bullet$ & $\bullet$ & $\bullet$ & $\bullet$ &  & $\bullet$ & $\bullet$\\
0.0113 & contain.aso:s & $\bullet$ & $\bullet$ & $\bullet$ & $\bullet$ & $\bullet$ & $\bullet$ & $\bullet$ & $\bullet$ &  & $\bullet$ & $\bullet$ & $\bullet$ & $\bullet$ & $\bullet$ & $\bullet$ &  & $\bullet$ & $\bullet$ &  & $\bullet$ & $\bullet$ &  & $\bullet$ & $\bullet$ & $\bullet$ & $\bullet$ & $\bullet$ &  & $\bullet$ & $\bullet$\\
0.0108 & become.as:s & $\bullet$ & $\bullet$ & $\bullet$ & $\bullet$ & $\bullet$ & $\bullet$ &  & $\bullet$ & $\bullet$ & $\bullet$ & $\bullet$ &  & $\bullet$ & $\bullet$ & $\bullet$ & $\bullet$ & $\bullet$ &  & $\bullet$ & $\bullet$ & $\bullet$ & $\bullet$ & $\bullet$ & $\bullet$ & $\bullet$ & $\bullet$ & $\bullet$ &  & $\bullet$ & $\bullet$\\
0.0099 & form.aso:o &  & $\bullet$ & $\bullet$ &  &  &  & $\bullet$ &  & $\bullet$ &  & $\bullet$ &  &  &  &  &  & $\bullet$ &  &  & $\bullet$ & $\bullet$ & $\bullet$ &  & $\bullet$ &  &  & $\bullet$ &  & $\bullet$ & $\bullet$\\
0.0086 & collapse.as:s &  &  & $\bullet$ & $\bullet$ &  &  & $\bullet$ & $\bullet$ & $\bullet$ &  & $\bullet$ &  & $\bullet$ &  & $\bullet$ &  &  &  &  & $\bullet$ & $\bullet$ &  &  &  & $\bullet$ &  & $\bullet$ & $\bullet$ & $\bullet$ & \\
0.0085 & create.aso:o &  & $\bullet$ & $\bullet$ &  & $\bullet$ &  & $\bullet$ &  &  & $\bullet$ & $\bullet$ &  & $\bullet$ & $\bullet$ &  & $\bullet$ & $\bullet$ & $\bullet$ & $\bullet$ & $\bullet$ & $\bullet$ & $\bullet$ & $\bullet$ & $\bullet$ & $\bullet$ & $\bullet$ & $\bullet$ &  & $\bullet$ & $\bullet$\\
0.0082 & provide.aso:s & $\bullet$ & $\bullet$ & $\bullet$ & $\bullet$ & $\bullet$ & $\bullet$ & $\bullet$ & $\bullet$ & $\bullet$ &  & $\bullet$ & $\bullet$ & $\bullet$ & $\bullet$ & $\bullet$ &  & $\bullet$ & $\bullet$ & $\bullet$ & $\bullet$ & $\bullet$ & $\bullet$ & $\bullet$ & $\bullet$ & $\bullet$ & $\bullet$ & $\bullet$ &  & $\bullet$ & $\bullet$\\
0.0082 & organize.aso:o & $\bullet$ &  &  & $\bullet$ & $\bullet$ &  & $\bullet$ &  &  &  &  & $\bullet$ &  &  & $\bullet$ &  & $\bullet$ &  &  & $\bullet$ & $\bullet$ &  & $\bullet$ &  & $\bullet$ &  & $\bullet$ &  & $\bullet$ & \\
0.0082 & offer.aso:s & $\bullet$ & $\bullet$ & $\bullet$ & $\bullet$ & $\bullet$ & $\bullet$ & $\bullet$ & $\bullet$ & $\bullet$ & $\bullet$ & $\bullet$ & $\bullet$ & $\bullet$ & $\bullet$ &  &  & $\bullet$ &  & $\bullet$ & $\bullet$ & $\bullet$ &  & $\bullet$ & $\bullet$ & $\bullet$ & $\bullet$ & $\bullet$ &  & $\bullet$ & \\
\hline\end{tabular}}
 \end{center}
  \caption{
    Class 19: ``locative action''. At the top are listed the 20 most probable nouns in the $p_{LC}(n|19)$ distribution and their probabilities, and at left are the 30 most probable verbs in the $p_{LC}(v|19)$ distribution. 19 is the class index. Those verb-noun pairs which were seen in the training data appear with a dot in the class matrix.  Verbs with suffix $.as:s$ indicate the subject slot of an active intransitive. Similarily $.aso:s$ denotes the subject slot of an active transitive, and $.aso:o$ denotes the object slot of an active transitive.} 
  \label{class19}
\end{figure*}

The aim of this paper is to use only minimal, but yet precise
information for lexical ambiguity resolution. We will show that good results are obtainable by employing a
simple and natural look-up in a probabilistic class-labeled
lexicon for disambiguation. The lexicon provides a probability
distribution on semantic selection-classes labeling the slots of verbal
subcategorization frames. Induction of distributions on frames and
class-labels is accomplished in an unsupervised manner by applying the
EM algorithm.
Disambiguation then is done by a simple look-up in the
probabilistic lexicon. We restrict our attention to a definition of
senses as alternative translations of source-words. Our approach
provides a very natural solution for such a  target-language
disambiguation task---look for the most frequent target-noun whose
semantics fits best with the semantics required by the target-verb. We
evaluated this simple method on a large number of real-world
translations and got results comparable to related approaches such
as that of Dagan and Itai~\shortcite{DaganItai:94} where much more selectional
information is used. 

\section{Lexicon Induction via EM-Based Clustering}
\label{SecLexInd}

\subsection{EM-Based Clustering}

For clustering, we used the method described in Rooth et al.
\shortcite{Rooth:99}. There classes are derived from
distributional data---a sample of pairs of verbs and nouns, gathered
by parsing an unannotated corpus and extracting the fillers of
grammatical relations. The semantically smoothed probability of a pair
$(v,n)$ is calculated in a latent class (LC) model as $p_{LC}(v,n) =
\sum_{c \in C} p_{LC}(c,v,n)$. The joint distribution is
defined by $p_{LC}(c,v,n) = p_{LC}(c) p_{LC}(v|c) p_{LC}(n|c)$.
By construction, conditioning of $v$ and $n$ on each other is
solely made through the classes $c$. The parameters $p_{LC}(c)$, $p_{LC}(v|c)$,
$p_{LC}(n|c)$ are estimated by a particularily simple version of the EM
algorithm for context-free models. 
Input to our clustering algorithm was a training corpus of 1,178,698
tokens (608,850 types) of verb-noun pairs participating in the
grammatical relations of intransitive and transitive verbs and their
subject- and object-fillers. Fig. \ref{class19} shows an induced
class from a model with 35 classes.
Induced classes often have a basis in lexical semantics; class 19 can
be interpreted as {\it locative}, involving location nouns ``room'',
``area'', and ``world'' and verbs as ``enter'' and ``cross''.

\subsection{Probabilistic Labeling with Latent Classes using EM-estimation}

To induce latent classes for the object slot of a fixed
transitive verb $v$, another statistical inference step was performed.
Given a latent class model
$p_{LC}(\cdot)$ for verb-noun pairs, and a sample $n_1, \dots, n_M$
of objects for a fixed transitive verb, 
we calculate the probability of an arbitrary object noun $n \in N$ by
\(
p(n) = \sum_{c \in C} p(c,n) = \sum_{c \in C} p(c) p_{LC}(n|c).
\)
This fine-tuning of the class parameters $p(c)$ to the sample of
objects for a fixed verb is formalized again as a simple instance
of the EM algorithm.
In an experiment with English data, we used a clustering model
with 35 classes. From the maximum probability
parses derived for the British National Corpus with the
head-lexicalized parser of Carroll and Rooth
\shortcite{CarrollRooth:98}, we extracted frequency tables for
transitive verb-noun pairs. These tables were used to induce a small
class-labeled lexicon (336 verbs).

\begin{figure}[htbp]

{\footnotesize
  \begin{tabular}{cc}
    \begin{tabular}{|lr|}
      \hline
      cross.aso:o 19 & 0.692 \\
      \hline
      \hline
      mind & 74.2 \\
      road & 30.3 \\
      line & 28.1 \\
      bridge & 27.5 \\
      room & 20.5 \\
      border & 17.8 \\
      boundary & 16.2 \\
      river & 14.6 \\
      street & 11.5 \\
      atlantic & 9.9 \\
      \hline
    \end{tabular}
    &
    \begin{tabular}{|lr|}
      \hline
      mobilize.aso:o 6 & 0.386 \\
      \hline
      \hline
      force & 2.00 \\
      people & 1.95 \\
      army & 1.46 \\
      sector & 0.90 \\
      society & 0.90 \\
      worker & 0.90 \\
      member & 0.88 \\
      company & 0.86 \\
      majority & 0.85 \\
      party & 0.80 \\
      \hline
    \end{tabular}
  \end{tabular}
  }
\caption{Estimated frequencies of the objects of the transitive verbs
  {\it cross} and {\it mobilize}}
\label{cross-mobilize}
\end{figure}

Fig.~\ref{cross-mobilize} shows the topmost parts of the lexical
entries for the transitive verbs \emph{cross} and \emph{mobilize}.
Class 19 is the most probable class-label for the object-slot of
\emph{cross} (probability 0.692); the objects of
\emph{mobilize} belong with probability 0.386 to class 16, which is
the most probable class for this slot. 
Fig.~\ref{cross-mobilize} shows for each verb the ten nouns $n$ with highest estimated
frequencies $f_c(n) = f(n)p(c|n)$, where $f(n)$ is the frequency of $n$
in the sample $n_1, \ldots, n_M$. For example, the
frequency of seeing \emph{mind} as object of \emph{cross} is estimated
as 74.2 times, and the most frequent object of \emph{mobilize} is
estimated to be \emph{force}.

\section{Disambiguation with Probabilistic Cluster-Based Lexicons}
\label{SecAlgo}

In the following, we will describe the simple and natural lexicon
look-up mechanism which is employed in our disambiguation approach. 
Consider Fig. \ref{examples} which shows two bilingual sentences
taken from our evaluation corpus (see Sect. \ref{SecEval}). The
source-words and their corresponding target-words are highlighted in
\textbf{bold face}. The correct translation of the source-noun (e.g.
\emph{Grenze}) as determined by the actual translators is replaced by
the set of alternative translations (e.g. $\{$ border, frontier,
boundary, limit, periphery, edge $\}$) as proposed by the word-to-word
dictionary of Fig. \ref{dict35} (see Sect. \ref{SecEval}).

\begin{figure}[ht]
\hrulefill

{\footnotesize
(ID 160867) {\it Es  gibt  einige  alte  Passvorschriften,  die
  besagen,  
dass man  einen  Pass  haben  muss,  wenn  man  die  {\bf Grenze}
\"{u}berschreitet}.  
There  are  some  old  provisions  regarding  passports  which  state
that  people  crossing the  {\bf \{border/ frontier/ boundary/
  limit/ periphery/ edge\}} should  have  their
passport  on  them.
\newline
(ID 201946)
{\it Es  gibt  schliesslich  keine  L\"{o}sung  ohne  die  Mobilisierung
der b\"{u}rgerlichen {\bf Gesellschaft} und  die  Solidarit\"{a}t
der Demokraten  in  der  ganzen  Welt.}  
There  can  be  no  solution, finally, unless  civilian 
{\bf \{company/ society/ companionship/ party/ associate\}}
is mobilized and  solidarity  demonstrated  by  democrats
throughout  the  world.
}

\hrulefill
\caption{Examples for target-word ambiguities}
\label{examples}
\end{figure}

The problem to be solved is to find a correct translation of the
source-word using only minimal contextual information. In our
approach, the decision between alternative target-nouns is done by
using only information provided by the governing target-verb. The key
idea is to back up this
minimal information with the condensed and precise information of a
probabilistic class-based lexicon. The criterion for choosing an alternative
target-noun is thus the best fit of the lexical and semantic
information of the target-noun to the semantics of the
argument-slot of the target-verb. This criterion is checked by a
simple lexicon look-up where the target-noun with highest estimated
class-based frequency is determined.
Formally, choose the target-noun $\hat n$ (and a class $\hat c$) such that 
\[
f_{\hat c}(\hat n) = \underset{n \in N, c \in
C}{\max}\; f_c(n)
\]
where $f_c(n)= f(n)p(c|n)$ is the estimated frequency of $n$ in the
sample of objects of a fixed target-verb, $p(c|n)$ is the
class-membership probability of $n$ in $c$ as determined by the
probabilistic lexicon, and $f(n)$ is the frequency of $n$ in the
combined sample of objects and translation alternatives\footnote{Note
that $p(\hat c) = \underset{c \in C}{\max}\; p(c)$ in most, but not
all cases.}. 

Consider example ID 160867 from Fig. \ref{examples}. The ambiguity to
be resolved concerns the direct objects of the verb $cross$ whose
lexical entry is partly shown in Fig. \ref{cross-mobilize}. Class 19 and the
noun \emph{border} is the pair yielding a higher estimated frequency
than any other combination of a class and an alternative translation
such as \emph{boundary}. Similarly, for example ID 301946, the pair of the target-noun
\emph{society} and class 6 gives 
highest estimated frequency of the objects of \emph{mobilize}.

\section{Evaluation}
\label{SecEval}

We evaluated our resolution methods on a pseudo-disambiguation task
similar to that used in Rooth et al. \shortcite{Rooth:99} for
evaluating clustering models.
We used a test set of 298 $(v,n,n')$ triples where 
$(v,n)$ is chosen randomly from a test corpus of pairs, and $n'$ is
chosen randomly according to the marginal noun distribution for the 
test corpus.
Precision was calculated as the number of times the 
disambiguation method decided for the non-random target noun
($\hat{n}=n$).

As shown in Fig.~\ref{pd-hsc}, we  obtained 88~\% precision for
the class-based lexicon (ProbLex), which is a gain of 9~\% over the best
clustering model and a gain of 15~\% over the human
baseline\footnote{Similar results for pseudo-disambiguation were
  obtained for a simpler approach which
  avoids another EM application for probabilistic class labeling. Here $\hat n$ (and
  $\hat c$) was chosen such that \(f_{\hat c}(v,\hat n) =
  \underset{c,n}{\max} (( f_{LC}(v,n) + 1 ) p_{LC}(c|v,n) ).\)
  However, the sensitivity to class-parameters
  was lost in this approach.}.

\begin{figure}[ht]
  \footnotesize{
  \begin{center}
    \begin{tabular}{|c|c|c|c|}
      \hline
      ambiguity
      &
      \begin{tabular}{c}
        human \\
        baseline
      \end{tabular}
      &        
      clustering
      &
      ProbLex
      \\
      \hline
      \hline
      2 & 73.5 \% & 79.0 \% & 88.3 \% 
      \\
      \hline
    \end{tabular}
    \caption{Evaluation on pseudo-disambiguation task for
      noun-ambiguity}
    \label{pd-hsc}
  \end{center}
  }
\end{figure}

The results of the pseudo-disambiguation could be confirmed in a
further evaluation on a large number of randomly selected examples of
a real-world bilingual corpus. The corpus consists of sentence-aligned
debates of the European parliament (mlcc = multilingual corpus for
cooperation) with ca. 9 million tokens for German and English.
From this corpus we prepared a gold standard as follows. We gathered word-to-word
translations from online-available dictionaries and eliminated German
nouns for which we could not find at least two English translations in
the mlcc-corpus. The resulting 35 word dictionary is shown in Fig.
\ref{dict35}. 
Based on this dictionary, we extracted all bilingual
sentence pairs from the corpus which included both the source-noun and
the target-noun. We restricted the resulting ca. 10,000 sentence pairs
to those which included a source-noun from this
dictionary in the object position of a verb. Furthermore, the target-object
was required to be included in our dictionary and had to appear in a
similar verb-object position as the source-object for an acceptable
English translation of the German verb. We marked the German noun
$n_g$ in the source-sentence, its English translation $n_e$ as
appearing in the corpus, and the English lexical verb $v_e$. For the
35 word dictionary of Fig. \ref{dict35} this semi-automatic procedure
resulted in a test corpus of 1,340 examples. The average ambiguity in
this test corpus is 8.63 translations per source-word. Furthermore, we
took the semantically most distant translations for 25 words which
occured with a certain frequency in the evaluation corpus. This gave a
corpus of 814 examples with an average ambiguity of 2.83 translations.
The entries belonging to this dictionary are highlighted in
\textbf{bold face} in Fig.~\ref{dict35}. The dictionaries and the
related test corpora are available on the web\footnote{\scriptsize \texttt{http://www.ims.uni-stuttgart.de/projekte/gramotron/}}.

\begin{figure*}[htbp]
\begin{center}
{\tiny
\begin{tabular}{|l||l|}
\hline
Angriff &aggression, assault, offence, onset,    onslaught,
 attack ,  charge,  raid,     whammy, inroad \\ 
{\bf Art} &  {\bf form, type, way}, fashion,  fit,  kind,     wise,  
   manner, species,  mode,  sort,     variety  \\
{\bf Aufgabe} & {\bf abandonment,  office, task}, 
    exercise,         lesson, giveup,   job ,   problem,  tax \\     
Auswahl & eligibility,      selection,        choice, varity,
assortment,       extract,  range,    sample \\  
Begriff & concept,  item,     notion,   idea \\   
{\bf Boden}   &  {\bf ground,   land,  soil},     floor,    bottom  \\
{\bf Einrichtung} & {\bf arrangement, institution}, constitution,     establishment,
feature,  installation, construction, setup, adjustment,
composition,\\
&      organization  \\   
{\bf Erweiterung} & {\bf amplification, extension},    enhancement,      expansion,
 dilatation, upgrading, add-on, increment \\       
{\bf Fehler} &  {\bf error, shortcoming},  
blemish,  blunder,  bug,      defect, demerit, failure,  fault, flaw, mistake,  trouble, slip, blooper,  lapsus \\ 
Genehmigung  &   permission,       approval,         consent,  acceptance,       approbation,      authorization \\  
{\bf Geschichte} & {\bf history,  story},    tale,     saga,     strip \\  
{\bf Gesellschaft} & {\bf company,    society}, companionship, 
 party,    associate \\
{\bf Grenze}  & {\bf border,       frontier}, boundary,          {\bf limit},    periphery, edge\\    
{\bf Grund} &  {\bf master,   matter,   reason}, base,     cause,    ground,     bottom  root \\    
{\bf Karte} &  {\bf card,     map},      ticket,   chart \\  
Lage  &  site,     situation,        position,         bearing,  layer,    tier \\   
Mangel &  deficiency, lack,     privation, want, shortage,
shortcoming,      absence,  dearth, demerit, desideratum, insufficiency, paucity,  scarceness \\
{\bf Menge}  & {\bf amount,   deal,   lot},      mass,     multitude,        plenty, quantity,
quiverful,        volume,  abundance, aplenty, \\
& assemblage , crowd,     batch,    crop,       heap, lashings, scores, set, loads,    bulk \\   
{\bf Pr\"{u}fung} &  {\bf examination,  scrutiny,   verification},
     ordeal,          test,
trial,    inspection, tryout,    \\
&   assay, canvass, check, inquiry,  perusal, reconsideration,
scruting \\            
{\bf Schwierigkeit} &  {\bf difficulty, trouble},       problem,  severity,
 ardousness,       heaviness \\
{\bf Seite} & {\bf page,    party,     side},      point,     aspect  \\   
{\bf Sicherheit}  & {\bf certainty, guarantee, safety},      immunity,
   security ,         collateral , doubtlessness, sureness,          deposit \\
{\bf Stimme} & {\bf voice,     vote},      tones \\ 
{\bf Termin}  & {\bf date, deadline,    meeting},     appointment,
time,      term \\        
{\bf Verbindung} & {\bf association, contact,  link},   chain,     conjunction,       connection, fusion,    
 joint , compound, alliance, catenation,        tie,       union,
 bond, \\
& interface,         liaison,    touch, relation, incorporation \\      
 Verbot &  ban,        interdiction,     prohibition, forbiddance \\     
{\bf Verpflichtung} &  {\bf commitment, obligation,  undertaking}, duty,
      indebtedness , onus, debt, engagement, liability,       bond \\     
{\bf Vertrauen}  &  {\bf confidence,  reliance,   trust},      faith,          assurance,        dependence,       private,  secret \\ 
{\bf Wahl}  &  {\bf election,         option},  choice ,   ballot,  
alternative, poll ,   list \\    
{\bf Weg}   &  {\bf    path,   road,     way},      alley,    route,  lane\\    
Widerstand  &    resistance,  opposition,
drag \\     
Zeichen & character,        icon,     sign,     signal,
symbol, mark,     token,    figure,   omen \\    
{\bf Ziel}  &  {\bf aim,    destination,  end},   designation,          target,  
goal, object,   objective,        sightings, intention,        prompt \\   
Zusammenhang &   coherence,        context,  contiguity,
connection \\      
{\bf Zustimmung}  &  {\bf agreement, approval,  assent},     accordance,            approbation, 
consent,  affirmation, allowance, compliance, compliancy,       acclamation\\
\hline
\end{tabular}
}
\end{center}
\caption{Dictionaries extracted from online resources}
\label{dict35}
\end{figure*}

We believe that an evaluation on these test corpora is a realistic simulation
of the hard task of target-language disambiguation in real-word
machine translation. The translation alternatives are selected from
online dictionaries, correct translations are determined as the
actual translations found in the bilingual corpus, no examples are
omitted, the average ambiguity is high, and the translations are
often very close to each other. In constrast to this, most other
evaluations are based on frequent uses of only two clearly distant senses
that were determined as interesting by the experimenters.

\def\precrec#1#2 {
  \begin{tabular}{c}
    P: #1 \% \\
    E: #2 \%
  \end{tabular}
}

\begin{figure*}[ht]
  {\footnotesize
  \begin{center}
    \begin{tabular}{|c|c|c|c|c|c|}
      \hline
      ambiguity 
      & 
      random 
      &
      \begin{tabular}{c}
        major \\
        sense
      \end{tabular}
      &
      \begin{tabular}{c}
        empirical \\
        distrib.
      \end{tabular}
      &
      clustering
      & ProbLex
      \\
      \hline
      \hline
      8.63 & 14.2 \% & 31.9 \% & \precrec{46.1}{36.2} & 43.3 \% &
      49.4 \% \\
      2.83 & 35.9 \% & 45.5 \% & \precrec{60.8}{49.4} & 61.5 \% & 68.2
      \% 
      \\
      \hline
    \end{tabular}
    \caption{Disambiguation results for clustering versus
      probabilistic lexicon methods}
    \label{clust-and-lex}
  \end{center}
  }
\end{figure*}

Fig. \ref{clust-and-lex} shows the results of lexical ambiguity
resolution with probabilistic lexica in comparison to simpler methods.
The rows show the results for evaluations on the two corpora with
average ambiguity of 8.63 and 2.83 respectively. Column 2 shows the
percentage of correct translations found by disambiguation by random
choice. Column 3 presents as another baseline disambiguation with the
major sense, i.e., always choose the most frequent target-noun as
translation of the source-noun. In column 4, the empirical distribution
of $(v,n)$ pairs in the training corpus extracted from the BNC is
used as disambiguator. Note that this method yields good results in
terms of precision (P = \#correct / \#correct + \#incorrect), but
is much worse in terms of effectiveness (E = \#correct / \#correct + \#incorrect + \#don't know). The reason for this is that even if the distribution of $(v,n)$ pairs is estimated quite precisely for the pairs in the
large training corpus, there are still many pairs which receive the
same or no positive probability at all. These effects can
be overcome by a clustering approach to disambiguation (column 5).
Here the class-smoothed probability of a $(v,n)$ pair is used to decide
between alternative target-nouns. Since the clustering model assigns a
more fine-grained probability to nearly every pair in its domain,
there are no don't know cases for comparable precision values. However, the
semantically smoothed probability of the clustering models is still
too coarse-grained when compared to a disambiguation with a
probabilistic lexicon. Here a further gain in precision and equally
effectiveness of ca. 7 \% is obtained on both corpora (column 6). We
conjecture that
this gain can be attributed to the combination of frequency
information of the nouns and the fine-tuned distribution on
the selection classes of the the nominal arguments of the verbs. We
believe that including the set of translation alternatives in the
ProbLex distribution is important for increasing efficiency, because it gives
the disambiguation model the opportunity to choose among unseen
alternatives. Furthermore, it seems that the higher precision of ProbLex can not
be attributed to filling in zeroes in the empirical distribution.
Rather, we speculate that ProbLex intelligently filters the empirical
distribution by reducing maximal counts for observations which do not
fit into classes. This might help in cases where the empirical
distribution has equal values for two alternatives.

\begin{figure}[htb]
{\footnotesize 
  \begin{center}
    \begin{tabular}{|l|l||c|c|}
      \hline
      source & target
      &
      correct
      &
      accept.
      \\
      \hline
      \hline
      Seite 
      & \begin{tabular}{l}
        page \\ side
      \end{tabular} 
      & 76.9 \%  & 76.9 \%  \\
      Sicherheit & 
      \begin{tabular}{l}
        guarantee \\ safety
      \end{tabular}
      & 93.8 \% &  93.0 \% \\
      Verbindung & 
      \begin{tabular}{l}
        connection \\ link
      \end{tabular}
      & 58.8 \% & 93.8 \% \\
      Verpflichtung &
      \begin{tabular}{l}
        commitment \\ obligation
      \end{tabular}
      & 73.2 \% & 94.1 \% \\
      Ziel & 
      \begin{tabular}{l}
        objective \\ target
      \end{tabular}
      & 72.5 \% & 85.5 \% \\
      \hline
      \hline
      overall precision & & 78 \% & 90 \% \\
      \hline
    \end{tabular}
    \caption{Precision for finding correct and acceptable translations by
      lexicon look-up}
    \label{best-acpt}
  \end{center}
  }
\end{figure}

Fig. \ref{best-acpt} shows the results for disambiguation with
probabilistic lexica for five sample words with two translations each.
For this dictionary, a test corpus of 219 sentences was extracted,
200 of which were additionally labeled with acceptable translations.
Precision is 78 \% for finding correct translations and 90 \% for
finding acceptable translations.

Furthermore, in a subset of 100 test items with average ambiguity 8.6,
a human judge having access only to the English verb and the set of
candidates for the target-noun, i.e. the information used by the
model, selected among translations. On this set, human precision was
39~\%.

\section{Discussion}
\label{SecDisc}

\begin{figure*}[ht]
  \footnotesize{
  \begin{center}
    \begin{tabular}{|l||c|c|c|c|c|c|}
      \hline
      \begin{tabular}{c}
        disambiguation \\
        method
      \end{tabular}
      &
      \begin{tabular}{c}
        corpus\\
        size
      \end{tabular} &
      ambiguity & 
      random & precision & 
      \begin{tabular}{c}
        random \\
        (standardized)
      \end{tabular}
      & 
      \begin{tabular}{c}
        precision \\
        (standardized)
      \end{tabular} 
      \\
      \hline
      \hline
      ProbLex  & 
      1 340 & 8.63 & 14.2 \% & 49.4 \% & 53.4 \% & 79.7 \% \\
      & 814 & 2.83 & 35.9 \% & 68.2 \% & 50.5 \% & 77.5 \% \\
      & 219 & 2    & 50.0 \% & 78.0 \% & 50.0 \% & 78.0 \% \\     
      \hline
      Dagan, Itai 94 &
      103 & 2.27 & 44.1 \% &  
      \begin{tabular}{c}
        P: 91.4 \% \\
        E: 62.1 \%
      \end{tabular}
      &
      50.0 \%
      &
      \begin{tabular}{c}
        P: 92.7 \% \\
        E: 66.8 \%
      \end{tabular}
      \\
      Resnik 97 &
      88 & 3.51 & 28.5 \% & 44.3 \% & 50.0 \% & 63.8 \% \\
      SENSEVAL 00 &
      2 756 & 9.17 & 10.9 \% &
      \begin{tabular}{c}
        P: 20-65 \% \\
        E: 3-54 \% 
      \end{tabular}
      &
      50.0 \%
      &
       \begin{tabular}{c}
        P: 60-87 \% \\
        E: 33-83 \% 
      \end{tabular}
      \\
      Yarowsky 95 & 
      $\sim$ 37 000 & 2 & 50.0 \% & 96.5 \% & 50.0 \% & 96.5 \% \\
      Sch\"{u}tze 92 & 
      $\sim$ 3 000 & 2  & 50.0 \% & 92.0 \% & 50.0 \% & 92.0 \% \\
      \hline      
    \end{tabular}
    \caption{Comparison of unsupervised lexical disambiguation methods.}      
    \label{compare}
  \end{center}
  }
\end{figure*}

Fig. \ref{compare} shows a comparison of our approach to
state-of-the-art unsupervised algorithms for word sense
disambiguation. Column 2 shows the number of test examples used to
evaluate the various approaches. The range is from ca. 100 examples to
ca. 37,000 examples. Our method was evaluated on test corpora of sizes
219, 814, and 1,340. Column 3 gives the average number of
senses/translations for the different disambiguation methods. Here the
range of the ambiguity rate is from 2 to about 9 senses\footnote{The
  ambiguity factor 2.27 attributed to Dagan and Itai's \shortcite{DaganItai:94} experiment is calculated
  by dividing their average of 3.27 alternative translations by their
  average of 1.44 correct translations. Furthermore, we calculated the
  ambiguity factor 3.51 for Resnik's \shortcite{Resnik:97} experiment
  from his random baseline 28.5~\% by taking $100/28.5$; reversely,
  Dagan and Itai's \shortcite{DaganItai:94} random baseline can be
  calculated as $100/2.27 = 44.05$. The ambiguity factor for \textsc{Senseval}
  is calculated for the noun task in the English \textsc{Senseval}
  test set.}. Column 4 shows the random baselines cited for the
respective experiments, ranging from ca. 11~\% to 50~\%. Precision values are given in
column 5. In order to compare these results which were computed for
different ambiguity factors, we standardized the measures to an
evaluation for binary ambiguity. This is achieved by calculating
$p^{1/\log_2 amb}$ for precision $p$ and ambiguity factor $amb$. The consistency of
this ``binarization'' can be seen by a standardization of the
different random baselines which yields a value of ca. 50~\% for all
approaches\footnote{Note that we are guaranteed to get exactly
  50~\% standardized random baseline if $random \cdot amb = 100 \; \%$.}. The standardized precision of our approach
is ca. 79~\% on all test corpora. 
The most direct point of comparison is the method
of Dagan and Itai \shortcite{DaganItai:94} which gives 91.4~\% precision (92.7
~\% standardized) and 62.1~\% effectiveness (66.8~\% standardized) on 103
test examples for target word selection in the transfer of Hebrew
to English. However, compensating this high precision measure for the
low effectiveness gives values comparable to our results.
Dagan and Itai's \shortcite{DaganItai:94} method is based on a large
variety of grammatical relations for verbal, nominal, and adjectival
predicates, but no class-based information or slot-labeling is used. 
Resnik \shortcite{Resnik:97}
presented a disambiguation method which yields 44.3~\% precision
(63.8~\% standardized) for a test set of 88 verb-object tokens. His approach
is comparable to ours in terms of informedness of the
disambiguator. He also uses a class-based selection measure, but based
on WordNet classes. However, the task of his evaluation was to select
WordNet-senses for the objects rather than the objects themselves, so the
results cannot be compared directly. The same is true for the
\textsc{Senseval} evaluation exercise \cite{Kilgarriff:00}---there
word senses from the \textsc{Hector}-dictionary had to be disambiguated. The
precision results for the ten unsupervised systems taking part in the
competitive evaluation ranged from 20-65\% at efficiency values from 3-54\%. 
The \textsc{Senseval} standard is clearly beaten by
the earlier results of Yarowsky \shortcite{Yarowsky:95} (96.5~\%
precision) and Sch\"utze \shortcite{Schuetze:92} (92~\% precision).
However, a comparison to these results is again somewhat difficult. Firstly,
these approaches were evaluated on words with two clearly
distant senses which were determined by the experimenters. In
contrast, our method was evalutated on randomly selected actual
translations of a large bilingual corpus. Furthermore, these
approaches use large amounts of information in terms of linguistic
categorizations, large context windows, or even manual intervention
such as initial sense seeding \cite{Yarowsky:95}. Such information is
easily obtainable, e.g., in IR applications, but often burdensome to
gather or simply unavailable in situations such
as incremental parsing or translation.

\section{Conclusion}
\label{SecConc}

The disambiguation method presented in this paper deliberately is
restricted to the limited amount of information provided by a
probabilistic class-based lexicon. This information yet proves itself
accurate enough to yield good empirical results, e.g., in
target-language disambiguation. The probabilistic class-based lexica
are induced in an unsupervised manner from large unannotated corpora.
Once the lexica are constructed, lexical ambiguity resolution can be
done by a simple lexicon look-up. In target-word selection, the most
frequent target-noun whose semantics fits best to the semantics of the
argument-slot of the target-verb is chosen. We evaluated our method on
randomly selected examples from real-world bilingual corpora which
constitutes a realistic hard task. Disambiguation based on
probabilistic lexica performed satisfactory for this task. The lesson
learned from our experimental results is that hybrid models combining
frequency information and class-based probabilities outperform both
pure frequency-based models and pure clustering models. Further improvements are to be
expected from extended lexica including, e.g., adjectival and
prepositional predicates.

\bibliographystyle{acl}

\end{document}